\documentclass{article}

\usepackage{microtype}
\usepackage{graphicx}
\usepackage{subcaption}
\usepackage{booktabs} % for professional tables

\usepackage[hyperfootnotes=false]{hyperref}

\usepackage[preprint]{icml2026}

\hypersetup{
  pdfsubject={Sparse autoencoder analysis of register and high-norm patch tokens in vision transformers}
}

\usepackage{amsmath}
\usepackage{amssymb}
\usepackage{mathtools}
\usepackage{amsthm}

\usepackage[capitalize,noabbrev]{cleveref}

\theoremstyle{plain}

\theoremstyle{definition}

\theoremstyle{remark}

\icmltitlerunning{Decoding Functional Roles of Register and Outlier Tokens}

\begin{document}

\twocolumn[
  \icmltitle{Decoding the Functional Roles of Register and High-Norm Patch Tokens in Vision Transformers}

  \begin{icmlauthorlist}
    \icmlauthor{Neel Varma}{}
    \icmlauthor{Andrew Rufail}{}
    \icmlauthor{Dipika Khullar}{}
    \icmlauthor{Vasu Sharma}{}
  \end{icmlauthorlist}

  \icmlkeywords{Vision Transformers, DINOv2, Register Tokens, High-Norm Patch Tokens, Sparse Autoencoders, Interpretability}

  \vskip 0.3in
]

\printAffiliationsAndNotice{}

\begin{abstract}
Self-supervised Vision Transformers (ViTs), such as DINOv2, learn rich visual representations, but the functions of their internal tokens remain poorly understood. Recent architectures introduce dedicated register tokens to reduce high-norm outlier patch tokens that emerge in background regions, yet the semantic and functional roles of both token types have not been fully established. In this paper, we analyze these roles by training sparse autoencoders (SAEs) on register-token and outlier-token activations in DINOv2. Using an automated interpretability pipeline, UMAP clustering, and CLIP-space cross-checks, we find that register-token features are more strongly associated with high-level semantic concepts. Outlier-token features, by contrast, are more often associated with lower-level structural, background, and texture-dominant patterns. Causal ablations further reveal a substantial functional asymmetry: disrupting top-activating register-derived features produces a 48.17\% drop in representation cosine similarity, whereas disrupting outlier-derived features produces only a 0.31\% drop. Together, our results provide evidence for token specialization in self-supervised ViTs.
\end{abstract}

\section{Introduction}
\begin{figure*}[t]
    \centering
    \begin{subfigure}[b]{0.48\textwidth}
        \centering
        \includegraphics[width=\linewidth]{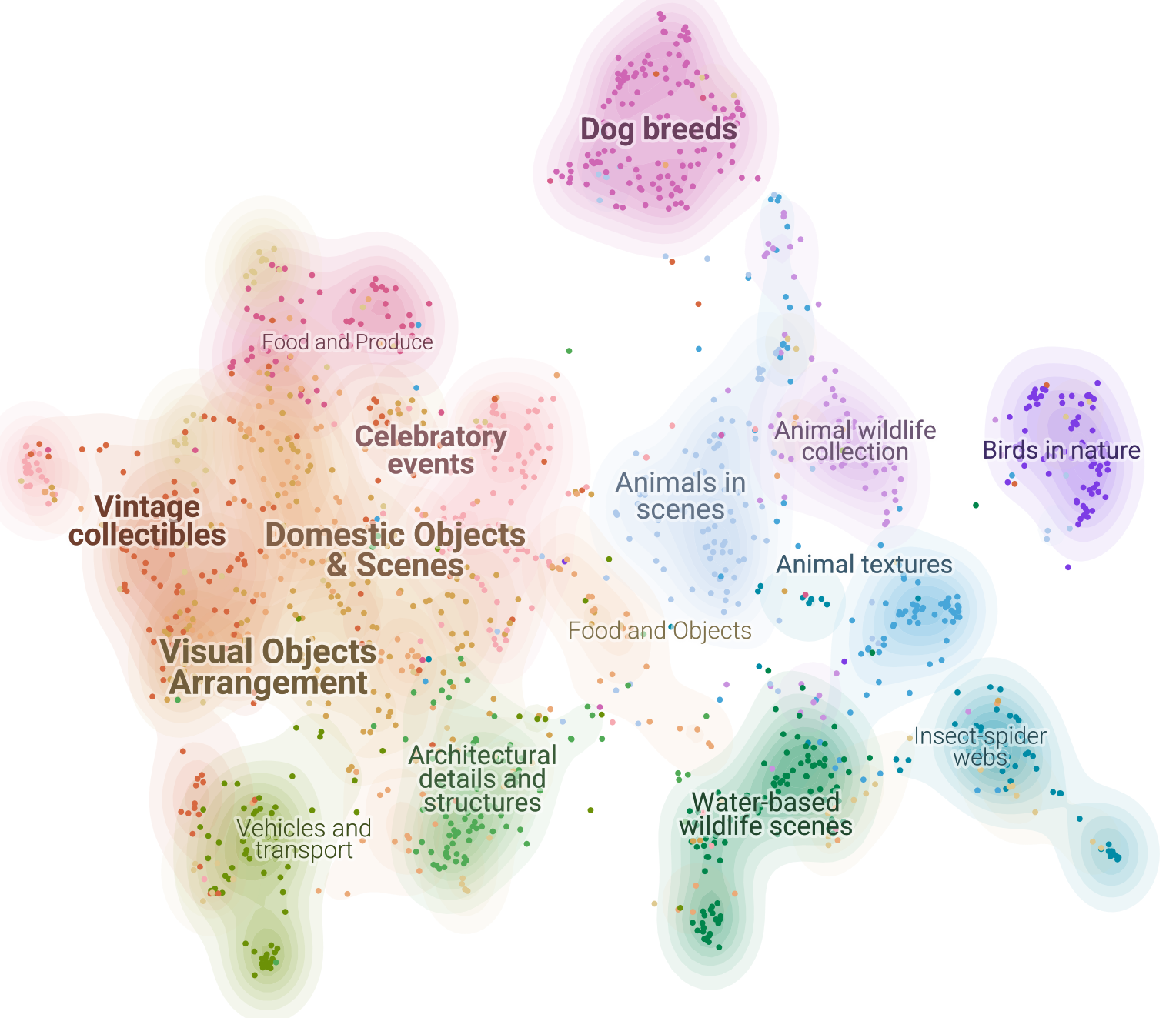}
        \caption{Register-only SAE features.}
        \label{fig:reg_only}
    \end{subfigure}
    \hfill
    \begin{subfigure}[b]{0.48\textwidth}
        \centering
        \includegraphics[width=\linewidth]{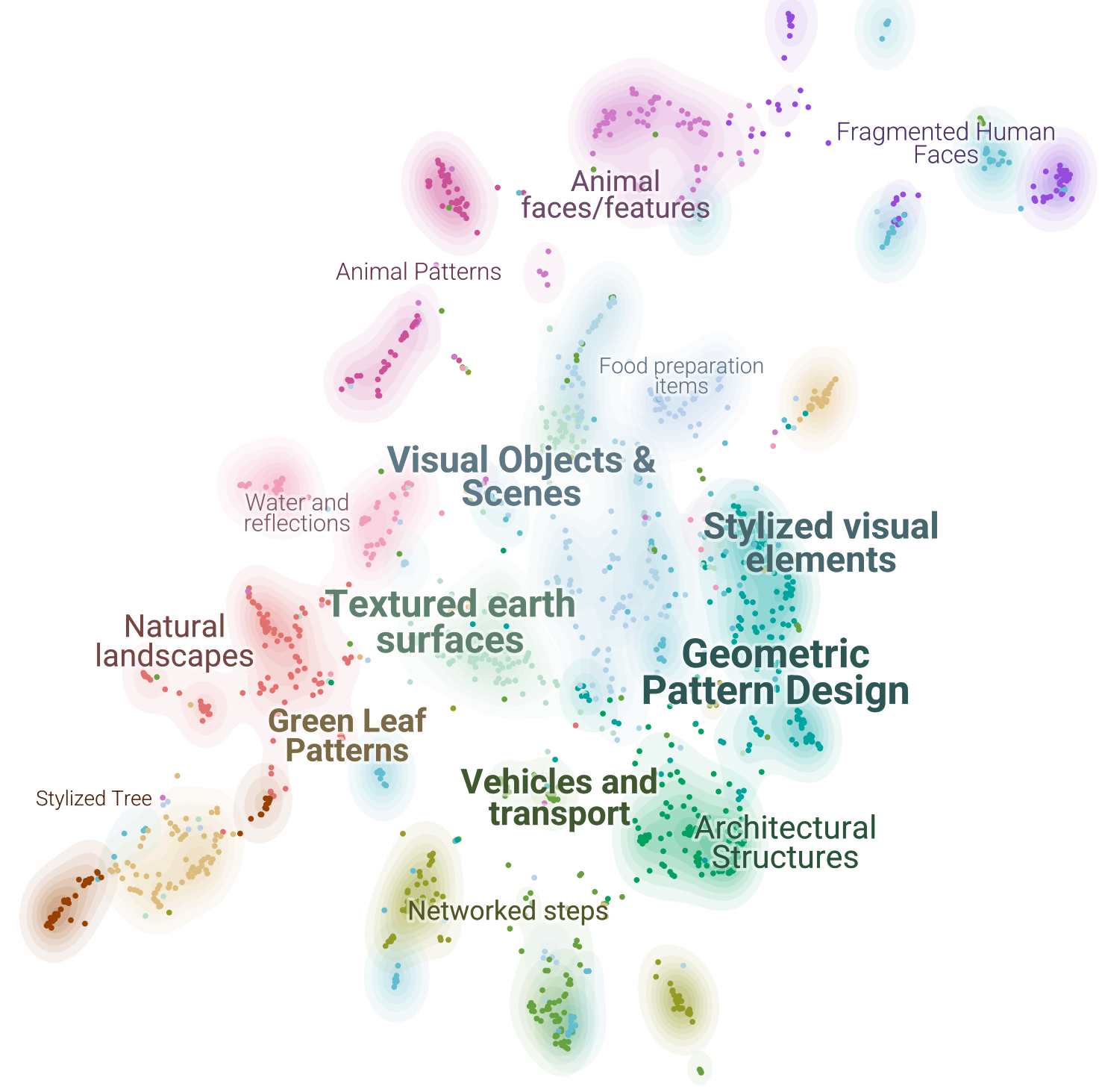}
        \caption{Outlier-only SAE features.}
        \label{fig:outlier_only}
    \end{subfigure}
    \caption{UMAP visualizations of the learned feature spaces. Register features (left) show semantic clustering, while outlier features (right) cluster around structural and texture-dominant information.}
    \label{fig:combined_umaps}
\end{figure*}

\begin{figure*}[t]
    \centering
    % Row 1: Bird Example
    \begin{subfigure}[b]{0.45\textwidth}
        \centering
        \includegraphics[width=\linewidth]{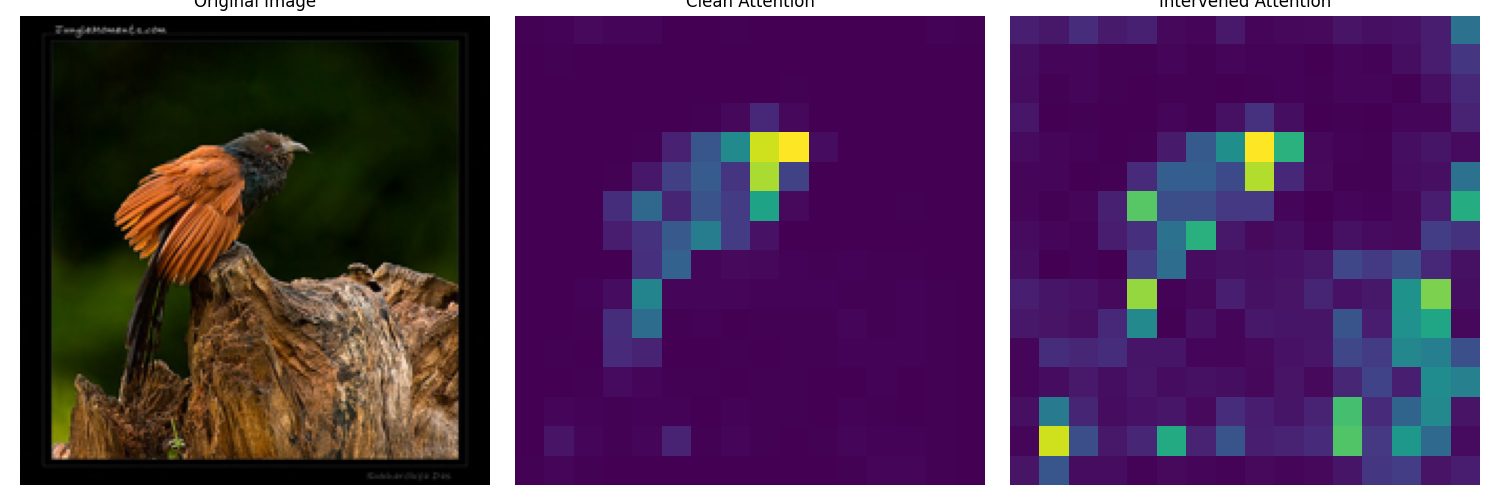}
        \caption{Bird: Attention Map}
        \label{fig:bird_attn}
    \end{subfigure}
    \hfill
    \begin{subfigure}[b]{0.45\textwidth}
        \centering
        \includegraphics[width=\linewidth]{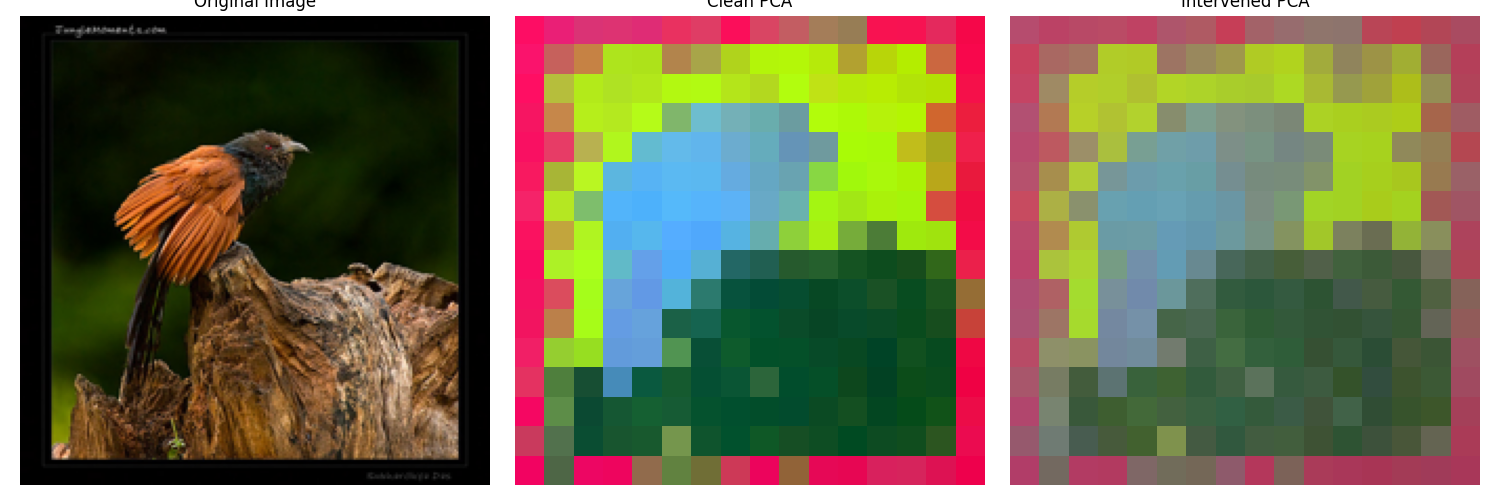}
        \caption{Bird: PCA Visualization}
        \label{fig:bird_pca}
    \end{subfigure}

    \vspace{1em} % Adds vertical space between the two rows

    % Row 2: Mountain Example
    \begin{subfigure}[b]{0.45\textwidth}
        \centering
        \includegraphics[width=\linewidth]{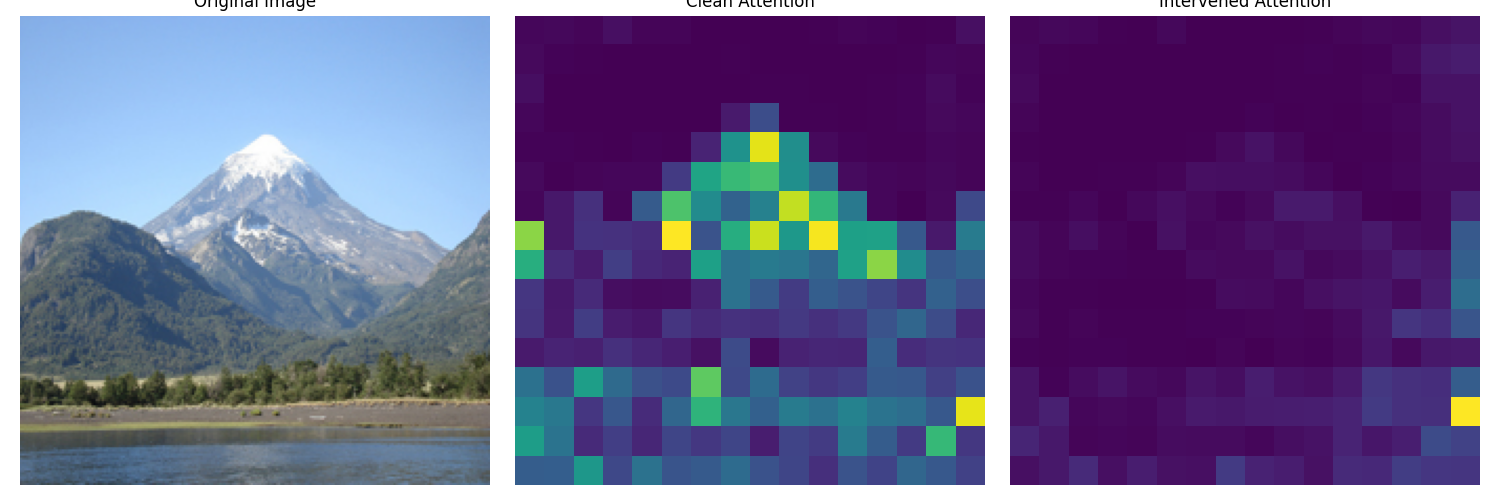}
        \caption{Mountain: Attention Map}
        \label{fig:mountain_attn}
    \end{subfigure}
    \hfill
    \begin{subfigure}[b]{0.45\textwidth}
        \centering
        \includegraphics[width=\linewidth]{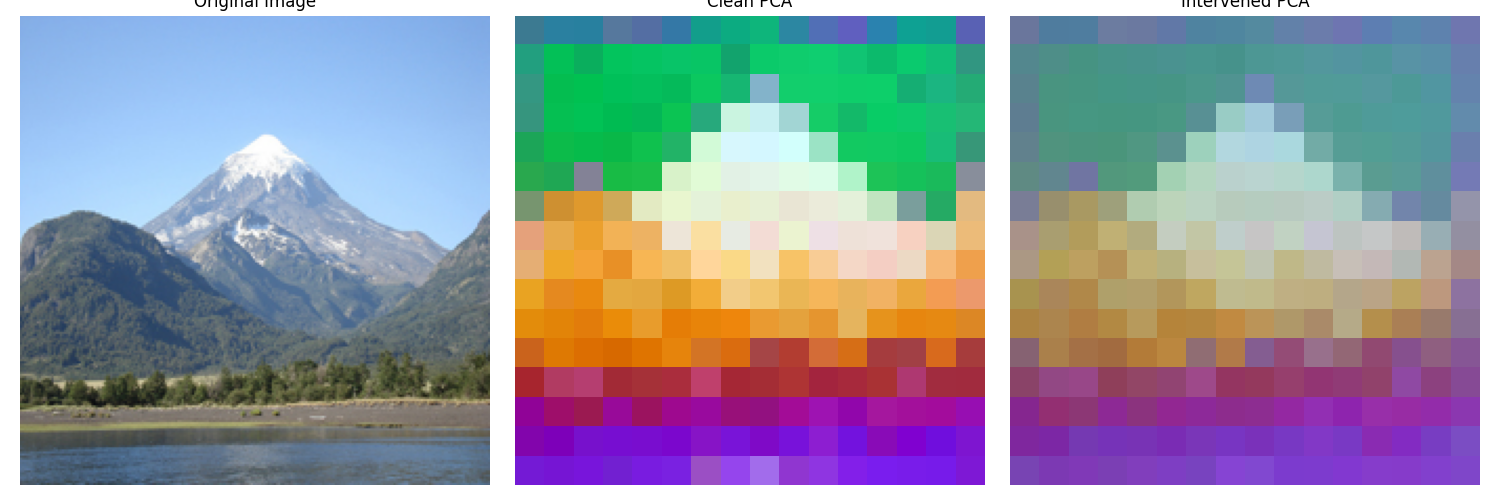}
        \caption{Mountain: PCA Visualization}
        \label{fig:mountain_pca}
    \end{subfigure}

    \caption{Visual impact of ablating the top-5 activating register features. The left column displays the resulting attention maps, while the right column shows PCA-based semantic maps where similar colors indicate similar semantic meanings. The interventions lead to significant disruption in object focus and semantic coherence.}
    \label{fig:ablation_results}
\end{figure*}

Despite the strong empirical performance of self-supervised Vision Transformers, the internal organization of their representations remains difficult to interpret, particularly when distinct token types may serve different functional roles.

Self-supervised Vision Transformers such as DINO and DINOv2 provide a useful setting for studying such token-level roles because their features encode meaningful spatial and semantic structure. Prior work has shown that these models support dense prediction and object discovery, suggesting that their internal feature maps carry rich information about image layout and content. At the same time, recent work has identified anomalous high-norm patch tokens, or outliers, that tend to appear in low-information background regions and may be repurposed for global computation rather than primarily encoding local spatial content.

Darcet et al. propose register tokens as a targeted architectural solution to this issue. By introducing dedicated non-image tokens into the input sequence, the model is given separate capacity for internal computation that would otherwise be absorbed by ordinary patch tokens. Their results show that register tokens reduce outlier artifacts, produce smoother feature and attention maps, and improve performance on dense prediction tasks.

However, an important interpretability question remains open. While prior work demonstrates that register tokens improve the spatial regularity of patch representations, it does not fully explain what information register tokens themselves encode, how their representations compare to high-norm outlier tokens, whether their feature spaces overlap, or whether they play distinct semantic and computational roles within the model.

To address these questions, we train sparse autoencoders on targeted token subsets from DINOv2, including register-only, outlier-only, and full-model activations. We interpret the learned features using top-activating examples, VLM-based descriptions, UMAP visualizations, and CLIP-space cross-checks, then assess functional importance through latent ablations that measure how feature interventions affect downstream patch representations and attention behavior.

Our results suggest that register-token features are more semantically coherent and more strongly aligned with high-level visual concepts, while outlier-token features are more often associated with structural or texture-dominant information. Ablations of register-derived features also produce substantially larger disruptions to representation stability, providing evidence that register and outlier tokens serve distinct functional roles within the model.

Our contributions are threefold. First, we introduce a sparse-feature analysis framework for comparing register and outlier-token representations in DINOv2. Second, we provide qualitative and causal evidence that register-token features are more semantically coherent and functionally important than outlier-token features. Third, we show that high-norm outlier tokens are more strongly associated with lower-level structural and texture-heavy information, positioning them as complementary rather than redundant components in the model’s information processing.

\section{Related Work}

Our work builds on prior research in self-supervised Vision Transformers, register-token architectures, sparse autoencoder-based interpretability, and automated feature explanation.

\textbf{Vision Transformers.} Self-supervised learning paradigms, particularly those based on DINO \citep{caron2021emerging} and DINOv2 \citep{oquab2023dinov2}, have shown that ViTs can learn rich semantic and spatial structure without explicit supervision. These models perform strongly on dense prediction tasks such as semantic segmentation and depth estimation, largely because their feature maps exhibit robust object-level alignment. However, as these models scale to the ``Huge'' and ``Giant'' DINOv2 regimes, prior work has observed the emergence of localized high-norm artifacts in both attention and feature maps \citep{darcet2023vision}.

\textbf{Outlier Tokens and Register Tokens.} High-norm outlier tokens form a small subset of patch tokens, approximately $2\%$ in DINOv2, and tend to appear in low-information or background regions while exhibiting disproportionately large $L_2$ norms. Prior work hypothesizes that these tokens are used to accumulate global information rather than encode local semantic content. To address this behavior, \citet{darcet2023vision} introduced register tokens as dedicated non-image tokens that provide additional capacity for internal or global computation, thereby reducing the need for patch tokens to carry high-norm signals and restoring greater spatial regularity. This modification was shown to improve performance on dense downstream tasks.

More recent work has argued that these artifacts arise as a consequence of the softmax bottleneck, in which a small number of tokens are driven to carry disproportionately large residual attention mass \citep{jiang2025vision}. Although prior studies have shown that such outliers can be identified and mitigated post hoc, their semantic content and relationship to register-based computation remain poorly understood.

\textbf{Sparse Autoencoders.} To study these representations, we build on recent work in sparse autoencoders (SAEs). SAEs decompose dense activations into a sparse set of more interpretable latent features, making them useful for studying polysemanticity and feature-level structure in neural networks \citep{cunningham2023sparse}. Although much of the prior SAE literature has focused on text representations \citep{gao2024scaling, templeton2024scaling}, their use in vision is growing. Recent studies, such as \citet{fel2025archetypal}, have identified monosemantic visual concepts ranging from low-level textures to high-level semantic objects in models such as CLIP and DINO. Our work extends this line of research by applying SAEs specifically to register and outlier tokens in order to study their respective semantic and functional roles.

\textbf{Automated Interpretability.} Recent work has also explored automated explanation methods for visual SAE features using vision-language models. In particular, \citet{ferrando2026language} propose a steering-based approach in which individual SAE features are causally injected into a vision encoder and then described by a language model. This line of work suggests that intervention-based explanations can complement traditional top-activating-image methods while reducing dependence on purely correlational evidence. Our work is related in its use of automated feature interpretation, but differs in that we focus on characterizing the semantic and causal roles of register and outlier tokens in DINOv2.

\section{Methodology}

\begin{figure*}[t]
      \centering
      \includegraphics[width=\textwidth]{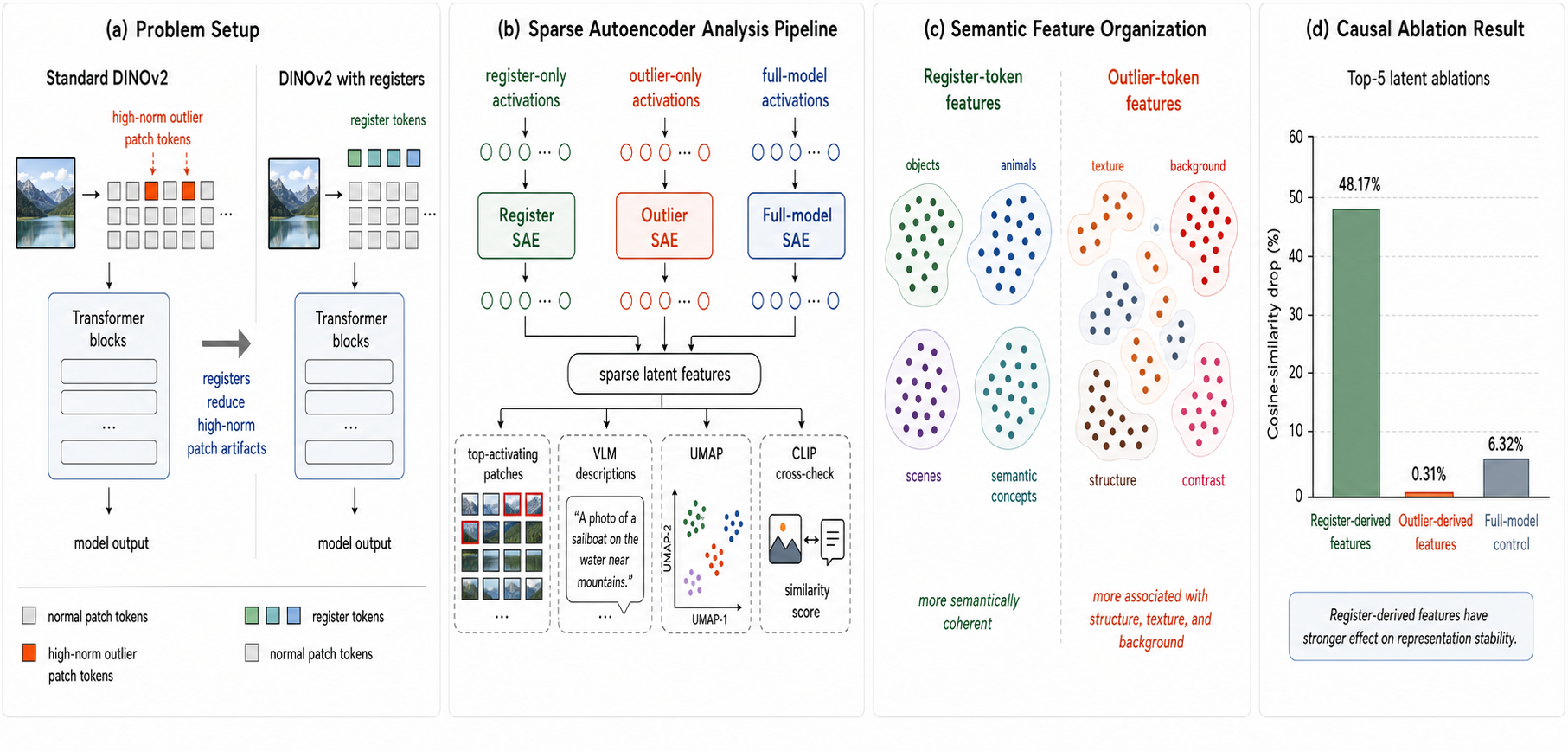}
      \caption{Overview of the register–outlier analysis pipeline. We isolate register tokens and high-norm outlier patch tokens, train token-specific sparse autoencoders, interpret learned features using top-activating patches, VLM captions, UMAP structure, and CLIP-space checks, and measure functional importance through latent ablations.}
      \label{fig:visualized-workflow}
  \end{figure*}
To disentangle the functional roles of register tokens and high-norm outlier patch tokens in DINOv2, we analyze intermediate activations from both a standard pretrained DINOv2 model and a register-augmented DINOv2 model. We partition layer-8 token activations into three groups: register tokens, high-norm outlier patch tokens, and standard patch tokens. We then train separate sparse autoencoders (SAEs) on each token subset, allowing us to decompose dense activations into sparse feature sets that can be compared across token types. Finally, we interpret the learned SAE features using top-activating image patches, Gemma 3-generated descriptions, UMAP visualizations, CLIP-space semantic checks, and causal feature ablations.

For the standard DINOv2 baseline, we use \textit{facebook/dinov2-small}; for the register-augmented model, we use \textit{facebook/dinov2-with-registers-small}. All SAEs are trained on layer-8 residual-stream activations extracted from ImageNet-1k images. Following Darcet et al., we define outlier tokens as high-norm patch tokens that tend to appear in low-information background regions. Following the outlier-token rate reported by Darcet et al., we select the top 2.37\% of layer-8 patch tokens by $L_2$ norm. The remaining patch tokens are treated as standard patch tokens.

Before SAE training, we normalize activations using a global scalar computed over the training activations for each token subset. Given activation vectors $x \in \mathbb{R}^d$, we compute
\[\sigma = \sqrt{\mathbb{E}[\|x\|_2^2]} .\]
We divide each activation by $\sigma$ before training so that the normalized activations have unit expected squared norm. This normalization improves hyperparameter stability and facilitates transferability across layers.

Each SAE encodes the normalized activation into a sparse latent representation:
\[f(x) = \operatorname{ReLU}\left(W_{\mathrm{enc}}(x - b_{\mathrm{dec}}) + b_{\mathrm{enc}}\right),\]
and reconstructs the activation through the decoder:
\[\hat{x} = W_{\mathrm{dec}} f(x) + b_{\mathrm{dec}} .\]
The models are optimized with a loss consisting of a mean-squared reconstruction term and an \(L_1\) sparsity penalty:
\[\mathcal{L} = \|x - \hat{x}\|_2^2 + \lambda \|f(x)\|_1 .\]
The reconstruction term encourages the SAE to preserve the original activation, while the \(L_1\) penalty encourages sparse feature activations.

We evaluate the quality of the learned features using two metrics:
\begin{enumerate}
    \item Sparsity ($L_0$): the average number of nonzero feature activations per token.
    \item Fraction of variance explained (FVE): the proportion of activation variance captured by the reconstruction,
    \[\mathrm{FVE} = 1 - \frac{\operatorname{Var}(x - \hat{x})}{\operatorname{Var}(x)} .\]
\end{enumerate}

\subsection{Automated Interpretability}

To interpret the learned SAE features, we identify the top \(k=9\) patch-token activations for each feature. We then retrieve the corresponding image patches and use Gemma 3 as a vision-language model to generate descriptive captions for each patch cluster. These captions summarize the visual primitives associated with each learned feature.

To analyze the global organization of the learned feature space, we apply UMAP to the SAE latent features. As an additional semantic cross-check, we project DINOv2 activations into CLIP space and measure alignment between SAE features and CLIP text concepts. This provides an independent reference point for evaluating whether register-token features are more semantically coherent than outlier-token features, and whether outlier-token features are more strongly associated with structural or texture-dominant information.

\subsection{Causal Intervention}

To assess the functional roles of different token types, we perform latent feature ablations in SAE space and measure their downstream effects on patch representations and attention behavior. For each input image, we compute the SAE latent representation \(f(x)\), identify the top five most active features, set the corresponding feature activations to zero, and decode the ablated representation to obtain \(x_{\mathrm{ablate}}\).

We quantify the impact of these interventions using two downstream analyses. First, we perform principal component analysis (PCA) on the final-layer patch tokens. By comparing the principal components of the clean and intervened representations, we measure the degree of distortion induced in the representation space. Second, we extract self-attention weights from the final transformer block and compute global attention maps by averaging attention across heads. We use these maps to evaluate whether ablating selected SAE features changes the model’s focus on the primary semantic object in the image.

\section{Results}

\subsection{SAE Training}

By training sparse autoencoders (SAEs) on isolated subsets of register and outlier activations, we decompose the activation space into sparse, interpretable features. We train the \textit{Outlier only} and \textit{Base full} SAEs on activations from the original DINOv2
model, while the \textit{Register only} and \textit{Register full} SAEs are trained on activations from the register-augmented DINOv2-small model.

\begin{figure*}[t]
    \centering
    \includegraphics[width=0.9\textwidth]{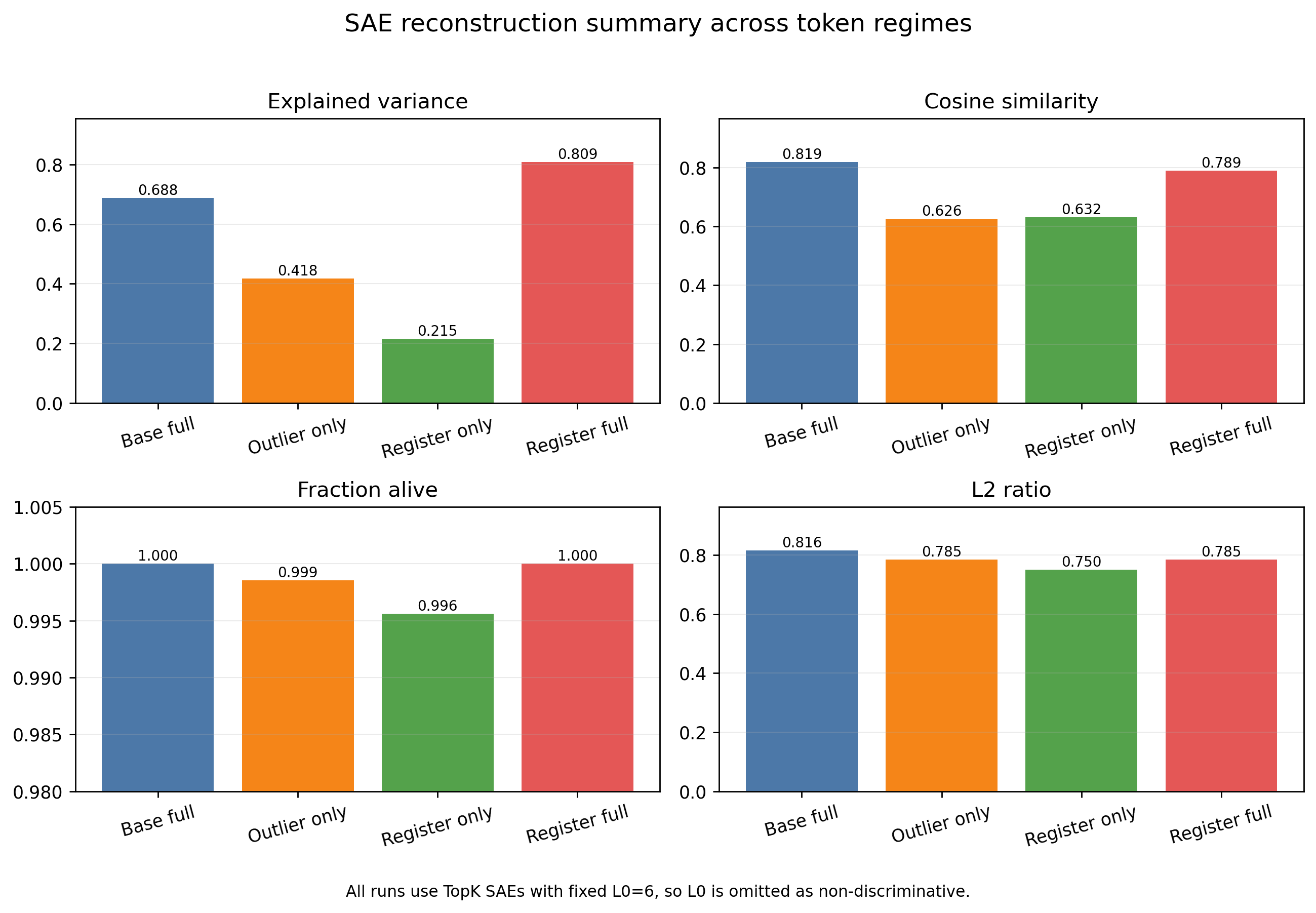}
        \caption{SAE training metrics and losses. Overall, all SAEs do not suffer from dead neurons. The full model SAEs have better reconstruction cosine similarities compared to outlier and register ones.}
        \label{fig:sae_losses}
\end{figure*}

The bar charts in Figure~\ref{fig:sae_losses} evaluate how well the SAEs reconstruct the original model activations. The \textit{Register full} regime achieves the highest explained variance (0.809), suggesting that including register tokens together with patch tokens makes the overall activation space easier for the SAE to model. Cosine similarity is highest for the \textit{Base full} SAE. In contrast, training only on specialized tokens produces reconstructions that are less aligned with the original activation vectors (\(\approx 0.63\)). Importantly, all models maintain nearly all of their features alive, indicating that they do not suffer from a substantial dead-neuron problem.

\begin{figure*}[t]
        \centering
        \includegraphics[width=\linewidth]{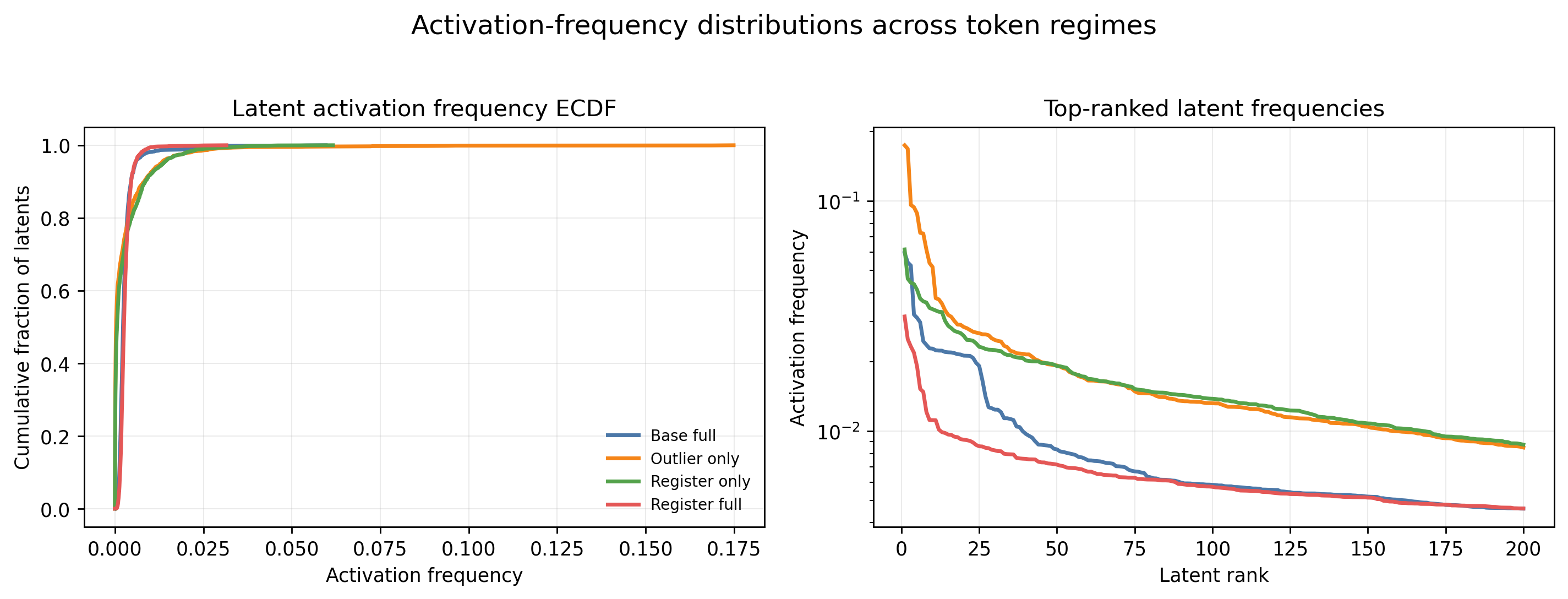}
        \caption{SAE latent activation-frequency distributions across token regimes. Most latents activate infrequently, indicating sparse feature usage across regimes.}
        \label{fig:freq_distributions}
\end{figure*}

Figure~\ref{fig:freq_distributions} shows how frequently SAE features are active. Across regimes, the vast majority of latents rarely fire, which is the intended behavior for sparse, interpretable models. However, the \textit{Outlier only} and \textit{Register only} curves extend farther to the right, indicating slightly denser feature activations than in full-token models.

\subsection{Clustering Activations}
To label the SAE features, we use Gemma 3 as a vision-language model (VLM) to generate captions for the top-activating patches associated with each feature.

Applying UMAP to the SAE latent space reveals a clear separation between token types. Features learned from register tokens cluster into high-level semantic categories, whereas features derived from outlier tokens show substantially lower semantic diversity and are concentrated around low-level background structure and texture.

A central finding of our analysis is what we term the \textit{texture sink} hypothesis. Our SAE-based interpretations suggest that high-norm outlier tokens, previously identified as anomalous artifacts in Vision Transformers \citep{darcet2023vision}, function as dedicated carriers of structural and textural visual information. Although these outlier tokens exhibit large \(L_2\) norms, VLM-based descriptions of their top-activating patches consistently emphasize background regions and texture-dominant content. This suggests that the model may use these tokens as a sink for structural information, limiting interference with semantic processing in the primary patch sequence.

To test whether these labels were merely artifacts of VLM bias, we projected the DINOv2 activations into CLIP space and re-examined the resulting feature organization. In this cross-modal embedding space, we observed strong alignment with the original SAE-VLM labels. This agreement provides additional evidence that register features encode higher-level semantic content, while outlier features act as structural sinks associated primarily with texture and background information.

\subsection{Causal Hierarchy}

To quantify the relative functional importance of different token types, we perform ablation studies targeting the top five activating SAE features across three settings: register-only features, outlier-only features, and a control condition using features from the full register-model SAE. For consistency with the token source, the outlier SAE is applied to the standard DINOv2 model, since register-augmented models largely suppress the high-norm outlier tokens observed in the standard model.

\begin{table}[h]
      \centering
        \begin{tabular}{l c c}
    \hline
    & \multicolumn{2}{c}{Cosine Similarity Drop} \\
    
    Ablation Target & Top-5 & Bottom-5 \\ \hline
    Register Tokens & 48.17\% & 47.05\% \\
    Outlier Tokens  & 0.31\%  & 0.28\%  \\
    Full Normal Model (control)     & 0.65\%  & 0.30\%  \\
    Full Register Model (control)     & 7.86\%  & 7.39\%  \\ \hline
\end{tabular}
      \caption{Reduction in cosine similarity after ablating the top-5 activating features in each condition. These causal ablations were performed on the model the SAE was trained on. The full models represent controls for their respective specialized tokens, both showing that the cosine similarity drops are significant.}
      \label{tab:cosine_sim}
  \end{table}

As shown in Table~\ref{tab:cosine_sim}, ablating the top five activating register features produces a 48.17\% reduction in representation stability. Attention maps show a corresponding degradation in the model's ability to localize the primary object in the image. In contrast, ablating outlier-derived features has a negligible effect, substantially smaller than the effect observed in the full-model control condition. This reveals a clear functional asymmetry: register-derived features are approximately 150 times more important for global representation stability than outlier-derived features. This finding is consistent with the UMAP structure, where outlier features cluster primarily around background and texture-based information.

Interestingly, ablating the bottom five activating register features produces a similarly large drop of 47.05\%. To investigate this mechanism, we perform iterative mean ablations over the register tokens, ablating one register token at a time up to all four registers. When only one register is mean-ablated, the cosine-similarity drop is negligible (\(\approx 0.01\)). However, ablating a second register produces an abrupt drop to \(\approx 0.50\), suggesting that the register system operates as a distributed buffer with a limited redundancy threshold. In DINOv2 with four default register tokens, the model appears to have roughly a one-register safety margin; once this capacity is exceeded, the texture-sink suppression mechanism fails and high-norm outlier-like signals reappear in the attention maps.

Our SAE analysis suggests that this failure mode is consistent across diverse image content. Once the redundancy threshold is exceeded, the top-activating latents identified by the SAE converge on a fixed set of indices, \(\{1153, 1454, 1554, 1389, 1460\}\), regardless of the input image. We term this pattern the \textit{Global Register Signature}. These findings provide a causal explanation for why top-five and bottom-five register ablations yield similar results: the catastrophic drop in similarity is not primarily determined by the semantic identity of the removed features, but by whether the intervention exceeds the sequestration capacity of the register-token system.

Thus, sufficiently strong disruption of register activations triggers a shared global failure mode. This suggests that the Global Register Signature does not correspond to image-specific semantic concepts, but instead reflects a universal stabilization or noise-sequestration function implemented by the register tokens. This supports the broader hypothesis that high-norm outlier signals are not primarily semantic representations, but are tied to structural or computational load that registers are designed to absorb.
\begin{figure}
      \centering
      \includegraphics[width=\columnwidth]{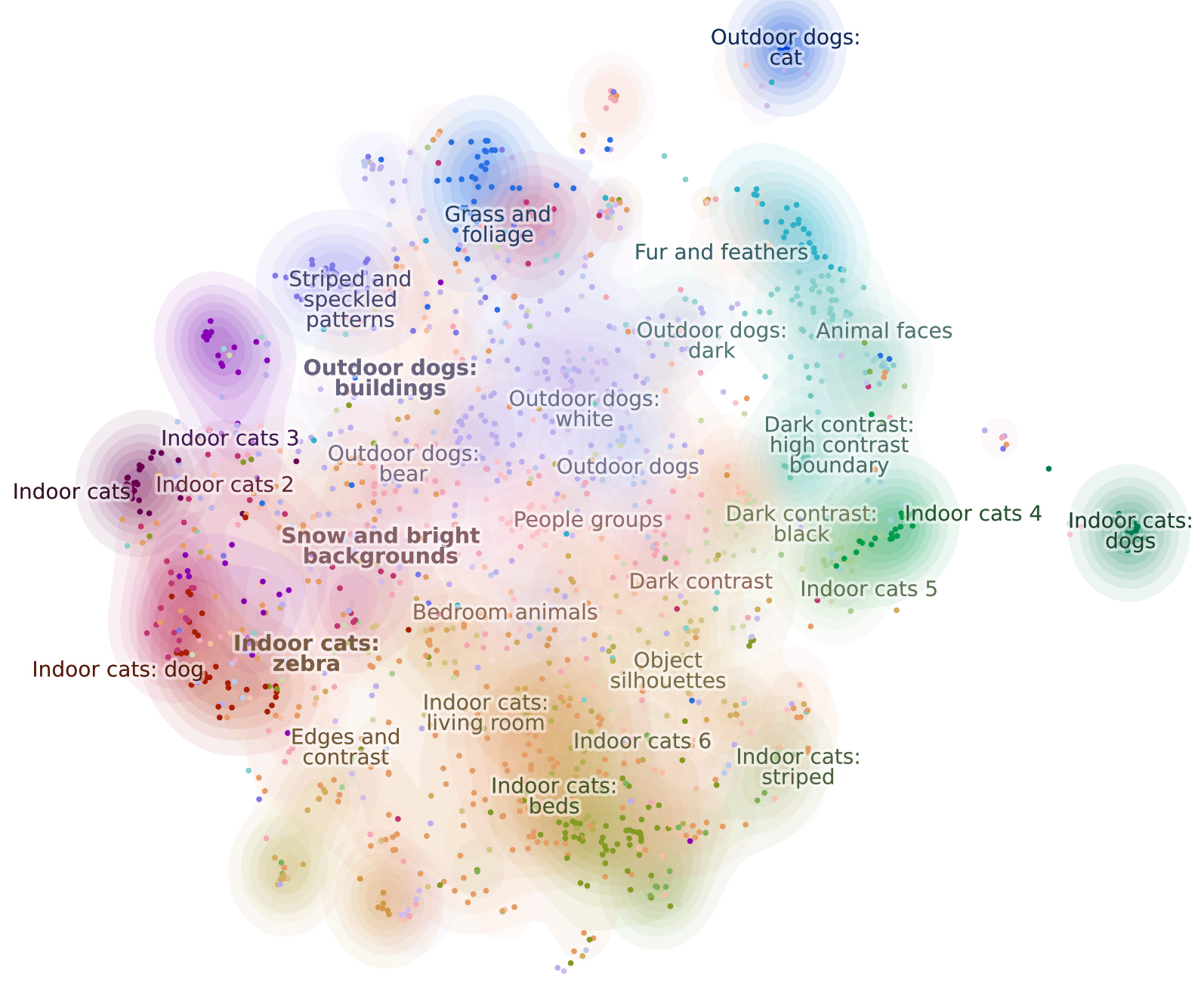}
      \caption{UMAP of outlier activations after projection into CLIP space. The feature organization remains consistent with
  the texture-sink interpretation.}
      \label{fig:clip_outlier}
  \end{figure}

\section{Conclusion}

In this study, we provide a mechanistic analysis of internal representations in self-supervised Vision Transformers, focusing on the relationship between register tokens in register-augmented DINOv2 and high-norm outlier patch tokens in standard DINOv2. By isolating targeted token subsets and training sparse autoencoders (SAEs), we obtain sparse feature dictionaries that allow these representations to be compared and interpreted.

Our analysis yields two main findings. First, register-derived features are more strongly associated with high-level semantic concepts, whereas standard-model outlier-derived features are more often associated with texture-dominant background structure. This supports the hypothesis that high-norm outliers are not merely anomalous artifacts, but may reflect a structured representational role tied to non-object-centered visual information. Second, causal latent interventions reveal a strong functional asymmetry: ablating top
 register-derived features substantially disrupts representation stability and downstream attention maps, while ablating outlier-derived features has a much smaller effect under the same protocol.

Together, these results suggest that DINOv2 variants separate semantic and structural information across distinct representational mechanisms. More broadly, our findings provide a foundation for analyzing, regularizing, and designing future ViT architectures with more interpretable token-level computation.

\subsection{Limitations}

The semantic and causal analyses reported here are limited to layer 8 of the DINOv2 Small checkpoints. Our accompanying artifact contains trained full-token, outlier-only, and register-only SAEs spanning DINOv2 Small, Base, and Large at layers 4, 8, and 11, but checkpoint availability alone does not establish that the reported semantic separation or causal asymmetry generalizes. Each checkpoint still requires the same top-activation annotation, quantitative semantic evaluation, and matched intervention protocol. In addition, the sparse autoencoders used in this study were trained under finite compute budgets, and stronger SAE training may yield cleaner or more stable feature dictionaries. Our interpretability pipeline also relies partly on qualitative tools, including top-activating examples, VLM-based descriptions, and UMAP visualizations. These tools are useful for exploratory analysis, but they do not by themselves establish a complete mechanistic account.

Finally, while our interventions provide evidence that register and high-norm patch tokens play different functional roles, they do not yet identify the circuit that routes information into these tokens or explain when that routing emerges during training.

\subsection{Future Work}

\begin{enumerate}
    \item \textbf{Replicate across model scales and layers.} We will repeat the semantic labeling, quantitative texture-versus-object scoring, and matched causal ablations at layers 4, 8, and 11 in DINOv2 Small, Base, and Large, both with and without registers. Because the accompanying artifact already contains 2,048-latent TopK SAE checkpoints for the full-token models and the token-specific outlier and register subsets across this grid, the remaining work is standardized analysis rather than SAE training. For every model--layer cell, we will report SAE reconstruction quality, semantic scores, and ablation effects with bootstrap confidence intervals, then test whether model scale or layer depth changes the register--outlier gap.

    \item \textbf{Trace the texture-routing circuit.} We will pair texture perturbations or high-norm source patches with content-matched controls and use activation patching to follow their influence through residual-stream states, attention heads, and MLP blocks. In register-augmented models, the endpoint will be activation of texture-associated register SAE features and the Global Register Signature; in standard models, it will be the corresponding high-norm outlier features. We will rank candidate edges by their mediated causal effect, then test the necessity and sufficiency of the resulting subgraph by ablating and restoring its highest-scoring edges. A stable patch-to-head/MLP-to-register or patch-to-head/MLP-to-outlier path would elevate the texture-sink hypothesis from a feature-level association to a circuit-level claim.

    \item \textbf{Identify when the behavior emerges during pretraining.} We will apply the same token-norm, SAE-feature, semantic, and causal-mediation measurements to chronological teacher checkpoints from an open DINO reproduction. A practical pilot is the public ImageNet-1k DINO ViT-S/16 reproduction with teacher and student revisions at epochs 100 and 300;\footnote{\url{https://huggingface.co/OK-AI/dino-vits16-pretrain-in1k}} a finer-grained follow-up can use the open DINOv2 training recipe, which periodically writes teacher snapshots.\footnote{\url{https://github.com/facebookresearch/dinov2}} We will keep images, token-selection percentiles, and evaluation code fixed across checkpoints and fit a change-point model to the outlier rate, texture score, and routing effect. Cross-reproduction differences will be treated as controls rather than as temporal evidence, so that any claimed onset is supported by checkpoints from a single training trajectory.
\end{enumerate}

\bibliography{example_paper}
\bibliographystyle{icml2026}

\end{document}